\pdfoutput=1
\def\year{2022}\relax
\documentclass[letterpaper]{article} 
\usepackage{aaai23}  
\usepackage{times}  
\usepackage{helvet}  
\usepackage[hyphens]{url}  
\usepackage{graphicx} 
\usepackage{natbib}  
\usepackage{caption} 
\usepackage{algorithm}
\usepackage{algorithmic}

\usepackage{newfloat}
\usepackage{listings}

\usepackage{amsmath}
\usepackage{diagbox}
\usepackage{subcaption}
\usepackage{bbding}

\title{\Large Just Noticeable Visual Redundancy Forecasting: A Deep Multimodal-driven Approach}
\author { 
    Wuyuan Xie,
    Shukang Wang,
    Sukun Tian,
    Lirong Huang,
    Ye Liu,
    Miaohui Wang \thanks{This work was supported in part by the National Natural Science Foundation of China under Grants 61701310 and 61902251, in part by the Natural Science Foundation of Shenzhen City under Grants 20200805200145001 and JCYJ20220809160139001, and in part by the Natural Science Foundation of Guangdong Province under 2021A1515011877 and 2022A1515011245. (\textit{Corresponding author}: \textit{Miaohui Wang})}
}

\begin{document}

\maketitle

\begin{abstract}
Just noticeable difference (JND) refers to the maximum visual change that   human eyes cannot perceive, and it has a wide range of applications in multimedia systems. However, most existing JND approaches only focus on a single modality, and rarely consider the complementary effects of multimodal  information. In this article, we investigate the JND modeling from an end-to-end homologous multimodal perspective, namely hmJND-Net. Specifically, we explore three important visually sensitive modalities, including saliency, depth, and segmentation. 
To better utilize homologous multimodal information, we establish an effective fusion method via summation enhancement and subtractive offset, and align homologous multimodal features based on a self-attention driven encoder-decoder paradigm. Extensive experimental results on eight different benchmark datasets validate the superiority of our hmJND-Net over eight representative methods.
\end{abstract}

\section{Introduction}
Just noticeable difference (JND) reflects the maximum visual redundancy (also known as \textit{visibility threshold}) that cannot be perceived by the human visual system (HVS). In multimedia community, visibility threshold effectively quantifies the perceptual redundancy in visual signals, which has a wide range of applications in image compression and processing systems \cite{wang2022perceptually}. For instance, JND can guide the removal of redundant information in multimedia compression \cite{zhang2016just, tian2021perceptual}, improving the coding efficiency without affecting its perceived quality. 
It can also improve the accuracy of quality assessment \cite{seo2020novel}, as well as increase the strength of watermark embedding \cite{liu2021360}. 
Based on the modeling approach, the existing JND methods can be roughly divided into 1) HVS-guided models \cite{bae2016hevc, wu2017enhanced, chen2019asymmetric} and 2) learning-based models \cite{huang2017measure, ki2018learning,  liu2019deep, shen2020just, tian2021perceptual}.
\begin{figure}[!t]
    \centering
    \includegraphics[width=0.99\linewidth, height=0.42\linewidth]{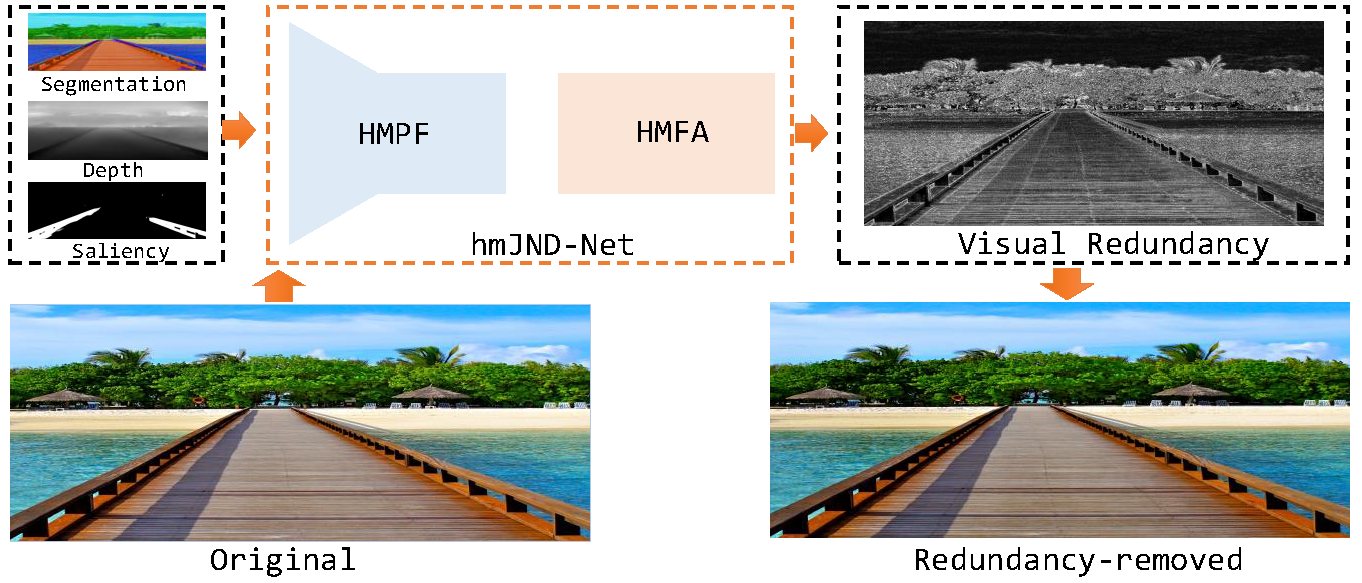}
    \caption{\textbf{Illustration of the proposed multimodal-driven visual redundancy forecasting network}. The saliency, depth, segmentation, and RGB modalities are jointly investigated to predict the visual redundancy based on a homologous-modality prior fusion (HMPF) module and a homologous-modality feature alignment (HMFA) module.}
    \label{fig_pipeline}
\end{figure}

The HVS-guided models mainly explore the visibility threshold  in the pixel or transform domain, both of which share a basic process \cite{liu2020novel}: first modeling some prominent visual masking effects, and then combining them with a  nonlinear fusion method. Based on the properties of HVS, the influence factors considered in existing methods are luminance adaptation (LA), spatial contrast masking (CM), visual saliency \cite{wang2014perceptual}, contrast sensitivity \cite{bae2016hevc}, pattern complexity \cite{wu2017enhanced},  and foveation effect \cite{chen2019asymmetric}, \textit{etc}. 
However, the HVS-guided approaches have at least two shortcomings: 1) Due to the fact that our knowledge for HVS is limited, we cannot obtain all possible influence factors to predict the just noticeable visual redundancy; 2) the potential interrelationship between different HVS factors cannot be accurately characterized by existing analytic models.

With the development of deep learning, some preliminary studies \cite{liu2019deep,shen2020just, tian2021perceptual} aim to address the problems of the HVS-guided model described above. Several encoder-based JND datasets have been constructed, which are essential for data-driven approaches. 
In labeling the visibility threshold map, each subject evaluates a series of distorted images or videos produced by a typical codec with different quality factors. For instance,  
\cite{jin2016statistical} constructed a  JPEG-compression-based dataset MCL-JCI, where each source image was encoded 100 times by the JPEG codec from 1 to 100, and the decoded images were evaluated by volunteers through a dichotomous search strategy to determine the picture-wise label data. 
\cite{shen2020just} further constructed a large dataset based on the versatile video coding (VVC) codec. 
In addition, \cite{wang2016mcl} constructed a video-wise dataset. 
However, the learning-based JND models rely heavily on manually labeled datasets, and face the problem of insufficient data. In view of this, we try to explore the training data by introducing homologous multimodalities to further improve the prediction accuracy of visual redundancy.

Inspired by the above discussion, we investigate an end-to-end homologous multimodal-driven visual redundancy forecasting network, which incorporates saliency, depth, segmentation, and RGB image modalities as shown in Figure~\ref{fig_pipeline}. 
More specifically, we firstly acquire depth, saliency, and segmentation modalities from the input data. Then, we fuse those prior multimodal features via a channel attention network,  and subsequently use a transformer network to align the RGB modality with the fused prior multimodal features to obtain the final visibility threshold.

The main contributions compared with the previous methods are summarized as follows. We for the first time investigate homologous multimodal information in a supervised visibility threshold modeling, and devise a novel end-to-end  homologous multimodal JND forecasting network (namely hmJND-Net). The proposed hmJND-Net alleviates the problem of insufficient training data as well as improves the prediction accuracy. To better utilize the three prior modalities including depth, saliency, and segmentation,  we further develop a new homologous-modality prior fusion (HMPF) module to early fuse these prior  features, and explore a homologous-modality feature alignment (HMFA)  module with a transformer structure to align RGB image modality and three prior modalities. Experimental results on eight benchmark datasets verify that our hmJND-Net outperforms representative schemes by providing higher accuracy and more bit-rate saving.

\section{RELATED WORK} \label{sec:related}
In this section, we first briefly review some representative just noticeable visual redundancy models, including HVS-guided and learning-based approaches. Then, we introduce the motivation of the proposed hmJND-Net.

\subsection{HVS-guided Methods}
As mentioned earlier, the basis for HVS-guided models is derived from the properties of the HVS. 
In the pixel domain, the LA and CM effects are widely-used: The LA effect responds to the different sensitivities of the HVS to visual signals with different background intensities, while the CM effect responds to small changes of the HVS to uniform or non-uniform content (\textit{i.e.}, visibility suppression). 
For instance, \cite{chou1995perceptually} proposed a classical JND framework that modeled LA as a quasi-parabolic model and derived CM from the local gradient intensity. 
Based on it,  \cite{yang2005just} developed a nonlinear additivity model for masking (NAMM) module to reduce the superposition effect between two masking effects. 
\cite{liu2010just} further decomposed the original image into structural and textural components to estimate edge masking and texture masking, which aimed to avoid the over-estimation of the edge and texture regions.  
\cite{wu2013just} established an auto-regressive model based on the free-energy principle to predict the order and disorder regions, and employed the disorder concealment effect in the visibility threshold estimation. 
Later, they utilized the diversity of orientations to measure the pattern complexity and derive the pattern masking \cite{wu2017enhanced}.
Recently, \cite{chen2019asymmetric} further introduced the foveation effect and the asymmetric visual sensitivity into the maximum visual redundancy modeling.
\cite{jiang2022towards} explored the characteristics of Karhunen–Lo{\`e}ve transform (KLT) coefficient to derive a perceptually lossless prediction.
\begin{figure}[!t]
\centering
\includegraphics[width=\linewidth]{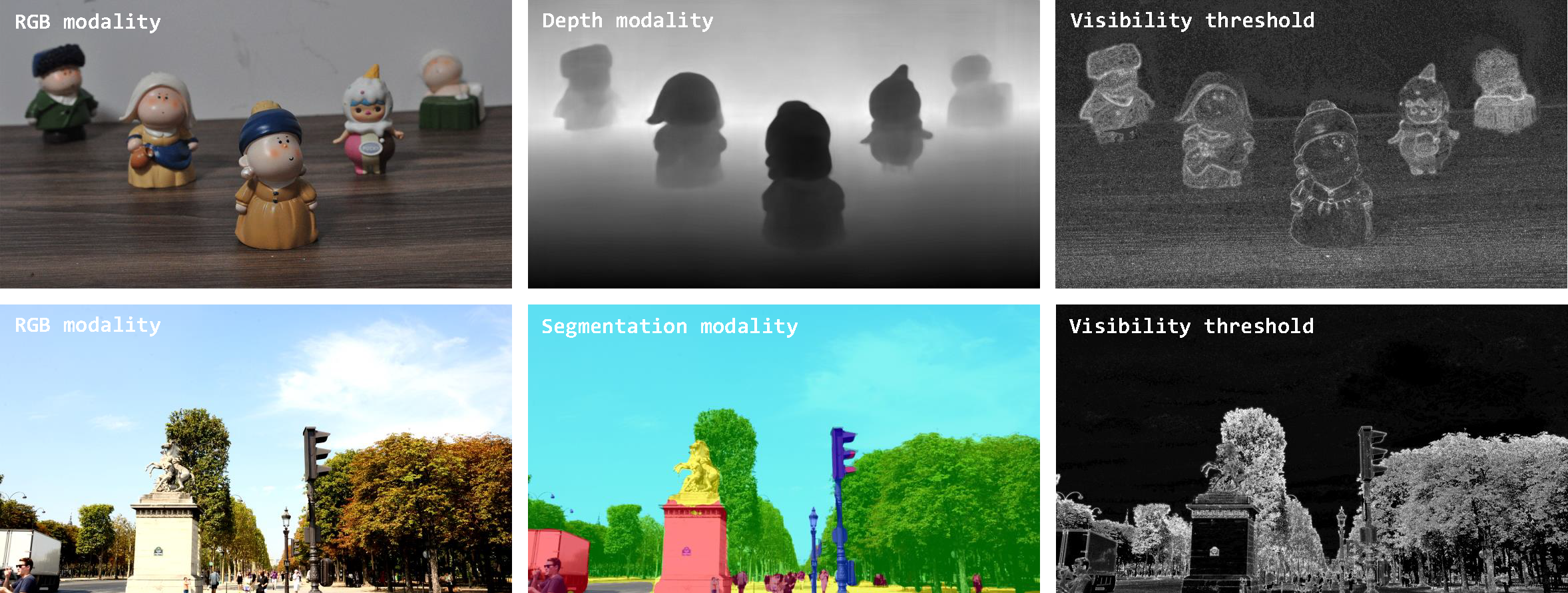}
\caption{\textbf{Examples of the depth modality and the segmentation modality}. 
}
\label{fig_motivation}
\end{figure}

Besides the pixel domain models,  the transform domain methods have  emerged to facilitate multimedia applications. The contrast sensitivity function (CSF) plays a dominant role in the transform domain JND modeling, which responds to the frequency characteristics of HVS. 
For instance, \cite{ahumada1992luminance} modeled the CSF component as an exponential function of spatial contrast. 
\cite{zhang2005improved} revised the LA and CM models based on a block classification in the discrete cosine transform (DCT) domain. 
\cite{wei2009spatio} used a gamma correction to compensate for the LA and CM effects in the DCT domain. 
\cite{bae2016hevc} established the CM effect based on the structural contrast of DCT blocks with multiple sizes, designed a DCT-based local redundancy probability to estimate the distribution of the transform coefficients, and introduced it into perceptual video coding.
\begin{figure*}[!t]
  \centering
  \setlength{\belowcaptionskip}{-0.4cm}
  \includegraphics[width=0.99\linewidth, height=0.25\linewidth]{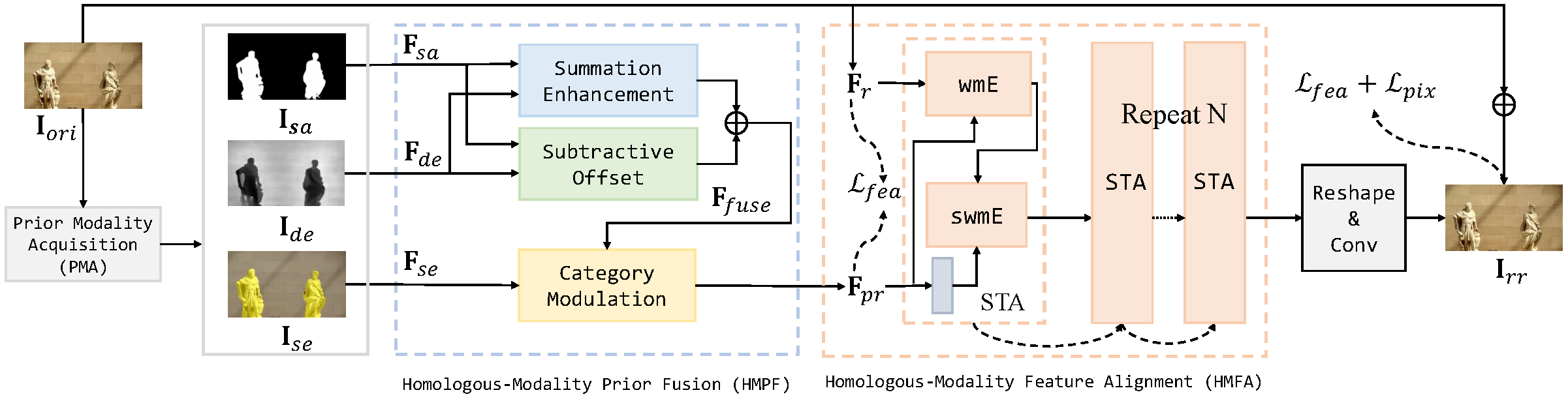}
  \hfill
  \caption{\textbf{Pipeline of the proposed multimodal-based visual redundancy framework}. It mainly consists of a prior modality acquisition (PMA) module, a homologous multimodal prior fusion (HMPF) module, and a homologous multimodal feature alignment (HMFA) module. \textbf{Please zoom in the electronic version for better details.}}
  \label{fig_overview}
\end{figure*}

\subsection{Learning-based Methods}
With the development of deep learning, learning-based JND estimation has become a new trend in this field. 
For instance,\cite{huang2017measure} proposed to learn the mean value of group-based visibility threshold distributions based on extracted video features. 
\cite{ki2018learning} established an energy reduction-based model, which used a linear regression and convolutional neural network (CNN)-based models to reduce the impact of the compressed artifacts.
\cite{liu2019deep} developed a deep binary classifier to predict whether an encoded image was perceptually distorted with respect to its original one. 
\cite{shen2020just} decomposed the input image into three components (\textit{i.e.}, luminance, contrast, and structure), and then trained a deep CNN degradation model to predict the visibility threshold.
\cite{tian2021perceptual} designed a block-level visual redundancy algorithm, which employed AlexNet to predict the encoding parameters for each image block. 
\cite{wang2022perceptually} predicted the visibility threshold by considering the oblique correction effect, where a  learned end-to-end CNN mapping was finally established between the visibility threshold and quality factor.


\subsection{Motivation}
Recently, multimodal data is usually used to portray an object from different sources or different perspectives. 
Data obtained from different sources refers to heterogeneous multimodal information \cite{yu2021ernie,  min2020study}. 
Data obtained from a single source refers to homologous multimodal information \cite{wang2018recovering,  shao2021generative, xie2022mnsrnet}. 
How to fuse homologous multimodal data is a popular research topic in multimedia image processing. 
Many studies have employed the homologous multimodal information to improve the performance of image processing systems. 
For example, \cite{wang2018recovering} introduced a segmentation modality to address the problem of unrealistic super-resolution textures. 
\cite{cheng2020zero} adopted a depth modality to extract the high and low resolution patches to train a super-resolution network. 
\cite{shao2021generative} utilized a saliency modality as the supervision for detail reconstruction in image inpainting tasks.

Inspired by the above efforts, we for the first time investigate the   homologous multimodal-based JND modeling (\textit{i.e.}, saliency, depth, segmentation) in this paper: 1) \textbf{Saliency modality.} Due to the fact that visual acuity decreases when the distance increases from the fovea of the human retina, visibility threshold increases with eccentricity. Thus, it is a natural choice to reduce less visual redundancy for salient regions and reduce more visual redundancy for non-salient regions. 
2) \textbf{Depth  modality.} The image patches close to the camera will be clearer, and as the depth is deeper, the image content will shrink and become blurred. As shown in Figure~\ref{fig_motivation}, there is a relationship between the visual redundancy and their depth, where a deeper depth tolerates more visual redundancy than the shallower one. Based on this prior knowledge, we can reduce more visual redundancy in the deeper depth regions based on the depth modality.
3) \textbf{Segmentation modality.} We find that different objects have various visibility thresholds. As shown in Figure~\ref{fig_motivation}, the visibility thresholds of trees and houses with complex textures are significantly higher than those of the flat sky and wall. Therefore, we introduce the category information into our framework via semantic segmentation modality, which reduces  different redundancies for  different objects.

\section{Proposed Multimodal Framework} \label{sec:method}
In this section, we describe the proposed end-to-end homologous multimodal  visibility threshold forecasting network in detail, which consists of a prior modality acquisition (PMA) module, homologous-modality prior fusion (HMPF) module, and homologous-modality feature alignment (HMFA) module as shown in Figure~\ref{fig_overview}.

\subsection{Overview}
\subsubsection{\textbf{Problem formulation}.} Unlike existing visibility threshold methods,  we perform a pixel-wise homologous multimodal network in an end-to-end manner, and hence our overall task can be expressed by minimizing a specially designed optimization problem as 
\begin{equation}
\mathop{\boldsymbol{\min}}\limits_{I_{rr}}\left(\mathcal{L}_{overall}\left(\mathcal{L}_{fea}(F_r,F_{pr}),\mathcal{L}_{pix}(I_{rr},I_{gt}\right)\right),
\end{equation}
where $\mathcal{L}_{overall}(\cdot)$ denotes the overall loss function.  $\mathcal{L}_{overall}(\cdot)$ is obtained by weighting the similarity measure $\mathcal{L}_{fea}$  between the RGB features $F_{{r}}$ and the prior fused features $F_{{pr}}$, and the pixel distance $\mathcal{L}_{pix}$  between the output redundancy-removed image $I_{rr}$ and its ground-truth $I_{gt}$.

More specifically,
\begin{equation}
\resizebox{.87\hsize}{!}{$
\begin{aligned}
&\mathcal{L}_{overall}\left(\mathcal{L}_{fea}(F_{{r}},F_{{pr}}),\mathcal{L}_{pix}(I_{rr},I_{gt})\right)\\ &=\lambda_{fea}\times \mathcal{L}_{fea}(F_{{r}},F_{{pr}}) + \lambda_{pix}\times \mathcal{L}_{pix}(I_{rr},I_{gt}) \\
&= \frac{\lambda_{fea}}{{h_{f} \times w_{f} \times c_{f}} } \|F_{{r}} - F_{{pr}}\|_1 + \frac{\lambda_{pix}}{{h_{p} \times w_{p} \times c_{p}} } \|I_{rr} - I_{gt}\|_2^2\\
\end{aligned}$},
\end{equation}
where ($h_{f}, w_{f}, c_{f}$) denotes the height, width, and the number of channels of the prior fused features. ($h_{p}, w_{p}, c_{p}$) denotes the height, width, and the number of channels of the predicted ones. 
$\mathcal{L}_{fea}$ denotes the pixel-wise $L_{1}$ loss between the RGB features and the prior fused features to measure their similarity, and minimizes  the distance between them for the multimodal feature alignment. 
$\mathcal{L}_{pix}$ denotes the pixel-wise  $L_{2}$ loss between the output redundancy-removed image and the ground-truth. $\lambda_{fea}$ and $\lambda_{pix}$ are the corresponding weights.

\subsubsection{\textbf{Architecture}.} We investigate a total of four modalities in the proposed hmJND-Net, including RGB image modality, saliency modality, depth modality, and segmentation modality.  
Figure~\ref{fig_overview} shows the overall network architecture,  which is formulated as
\begin{equation}
I_{rr} = \mathcal{M}_{JND}\left(I_{ori},\mathcal{M}_{EX}(I_{{hm}})\right),
\label{equ_overview}
\end{equation}
where $I_{{hm}}$ denotes the {homologous multimodal} information obtained via the original RGB modality $I_{ori}$, including saliency modality $I_{{sa}}$, depth modality $I_{{de}}$, and segmentation modality $I_{{se}}$. 
Generally, Eq.~(\ref{equ_overview}) can be decomposed into two sub-tasks: a  multimodal feature extraction module $\mathcal{M}_{EX}$ and a multimodal forecasting module $\mathcal{M}_{JND}$.

The multimodal feature extraction, $\mathcal{M}_{EX}$, consists of the PMA and {HMPF} modules. First, the original RGB image modality is fed into the PMA module to generate the corresponding three prior modalities, and the shallow features are extracted by a three-layer convolution. Then the shallow features are fed into the HMPF module, $\mathcal{M}_{{HMPF}}$, for the fusion of the prior modality features. 
$\mathcal{M}_{EX}(\cdot)$ can be expressed as
\begin{equation}
F_{{pr}} = \mathcal{M}_{HMPF}\left(\mathcal{M}_{PMA}(I_{ori})\right),
\label{equ_FCM}
\end{equation}
where $F_{{pr}}$ represents the prior fused features after a {homologous multimodal} fusion module, $\mathcal{M}_{PMA}(\cdot)$ denotes a PMA module, and $\mathcal{M}_{{HMPF}}(\cdot)$ denotes a HMPF module.

After the HMPF process,  $F_{{pr}}$ and $F_{{r}}$ are fed together into the HMFA module for the feature alignment. 
First, the pixel-wise distance between $F_{{pr}}$ and $F_{{r}}$ is  minimized, and then several Swin Transformer Alignment (STA) blocks are used to further align the $F_{{pr}}$ and $F_{{r}}$ features. Subsequently, the aligned features are fed into a \texttt{Reshape} module. Finally, the initial input is connected for the residual learning. $\mathcal{M}_{JND}(\cdot)$ can be expressed as
\begin{equation}
I_{rr} = \theta\left(\mathcal{M}_{RE}(\mathcal{M}_{{HMFA}}(F_{{pr}},F_{r}))\right),
\label{equ_MJND}
\end{equation}
where $F_{{r}}$ denotes the intermediate features of the RGB image modality  extracted by a three-layer convolution, $\mathcal{M}_{{HMFA}}(\cdot)$ denotes a HMFA module, $\mathcal{M}_{RE}(\cdot)$ denotes a \texttt{Reshape} module, and $\theta(\cdot)$ denotes the normalization operation to restrict the output to a reasonable scale.
\begin{figure}[!t]
    \centering
    \includegraphics[width=0.99\linewidth, height=0.5\linewidth]{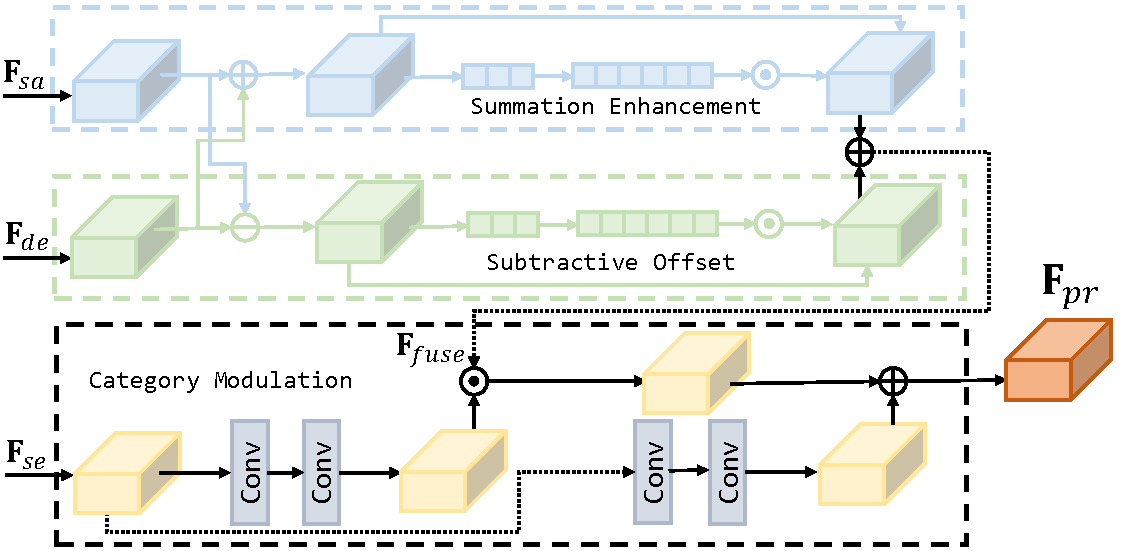}
    \captionsetup{font={small}}
    \caption{\textbf{Homologous-modality Prior Fusion (HMPF)}. The inputs are saliency $\textbf{F}_{sa}$, depth $\textbf{F}_{de}$ and segmentation $\textbf{F}_{se}$ modalities, and the output is the prior fused feature $\textbf{F}_{pr}$.}
    \label{fig_HMPF}
\end{figure}

{\subsection{Homologous-modality Prior Fusion}}
In order to make full use of the prior multimodal information, we develop a special homologous-modality prior fusion (HMPF) module as shown in Figure~\ref{fig_HMPF}. 
Before the HMPF process, the acquired prior modalities will be extracted by a three-layer 3$\times$3 convolution to obtain the corresponding shallow feature $F_{x}(i.e.,x={\left\{ {sa, de, se} \right\}})$. 
In other words, $I_{{sa}}$, $I_{{de}}$, and $I_{{se}}$ will be embedded into $F_{{sa}}$, $F_{{de}}$, and $F_{{se}}$, respectively. 
Given that the depth and saliency modalities on the visibility threshold estimation are not always positively correlated, image regions with a deeper depth may belong to salient, while those with shallower depth may belong to non-salient.
Therefore, to capture the {homologous multimodal} complementary information, {hmJND-Net} needs to learn not only the consistency between the depth and saliency modalities but also the difference between them.

Based on the above analysis, we propose a two-submodules strategy, summation enhancement $\mathcal{M}_{SE}$ and subtractive offset $\mathcal{M}_{SO}$, to learn the complementary relationship between these two modalities, and then modulate the fused feature map $F_{{fuse}}$ by introducing the category information. 
Specifically, we obtain the subtractive feature $F_{sub}$ with the difference information by subtracting $F_{{sa}}$ and $F_{{de}}$, and the summation feature $F_{sum}$ with the consistency information by adding $F_{{sa}}$ and $F_{{de}}$, respectively.

Inspired by the channel attention network SENet \cite{hu2018squeeze}, we further perform the channel attention weighting on these two intermediate features to obtain the enhanced feature $F_{{e}}$ and the offsetting feature $F_{{o}}$, and then fuse them to obtain $F_{{fuse}}$, which can be formulated as
\begin{equation}
F_{{fuse}} = F_{{e}} \oplus F_{{o}},
\label{equ_FUSE}
\end{equation}
\begin{equation}
F_{{e}} = F_{sum} \otimes \mathcal{F}_{{fc}}^{{s}}(\mathcal{F}_{{fc}}^{{r}}(\mathcal{F}_{{gp}}(F_{sum}))) \oplus F_{sum},
\label{equ_FE}
\end{equation}
\begin{equation}
F_{{o}} = F_{sub} \otimes \mathcal{F}_{{fc}}^{{s}}(\mathcal{F}_{{fc}}^{{r}}(\mathcal{F}_{{gp}}(F_{sub})))  \oplus F_{sub},
\label{equ_FO}
\end{equation}
where $\oplus$ and $\otimes$ denote an element-wise summation and multiplication, respectively. $\mathcal{F}_{{gp}}(\cdot)$ denotes a global pooling, and $\mathcal{F}_{{fc}}^{x}(\cdot)$ denotes a fully-connected layer followed by a \texttt{Sigmoid} normalization or a \texttt{ReLU} activation layer.

After obtaining the fusion features of saliency and depth modalities, we need to incorporate the category information. Inspired by the spatial feature transform \cite{wang2018recovering}, we propose to use the semantic segmentation modality,  and modulate the fused features based on it. This fusion process can be formulated as
\begin{equation}
F_{{pr}} = F_{{fuse}} \otimes \mathcal{F}_{conv}(F_{{se}}) \oplus \mathcal{F}_{conv}(F_{{se}}),
\label{equ_CM}
\end{equation}
where $\mathcal{F}_{conv}(\cdot)$ denotes a two-layer convolution. The fused features of all three modalities will be fed into the next {HMFA} module.
\begin{figure}[!t]
    \centering
    \includegraphics[width=0.99\linewidth, height=0.5\linewidth]{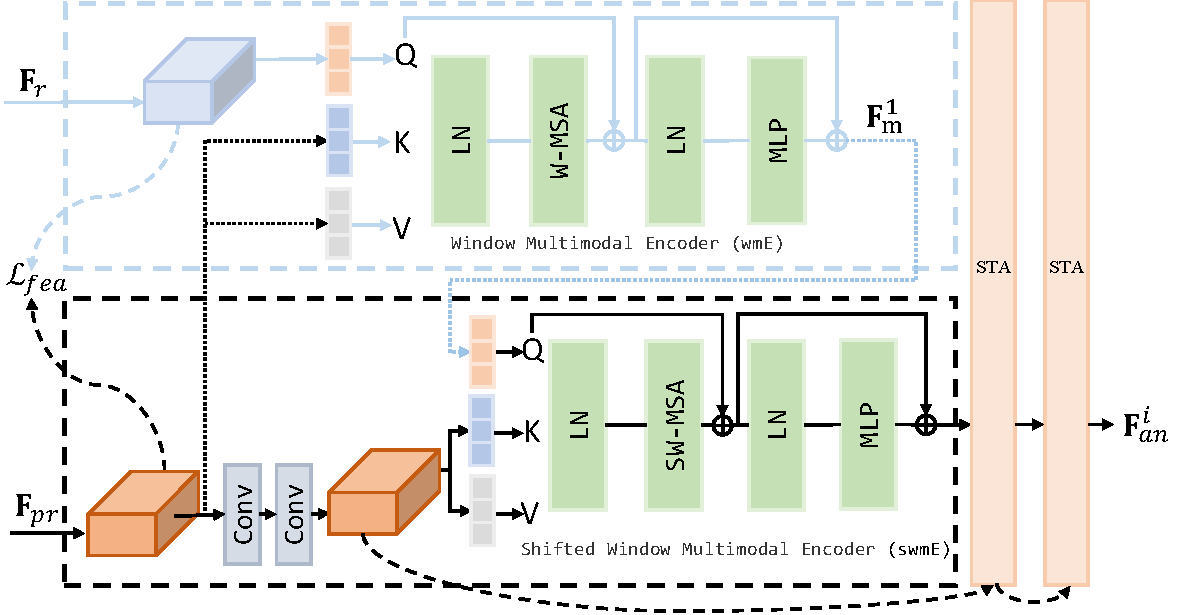}
    \captionsetup{font={small}}
    \caption{{\textbf{Homologous-modality Feature Alignment (HMFA)}. The inputs are the prior fused feature $\textbf{F}_{pr}$ and RGB modality feature $\textbf{F}_{r}$, and the output is the aligned feature $\textbf{F}_{an}^i$.
}}
    \label{fig_HMFA}
\end{figure}

{\subsection{Homologous-modality Feature Alignment}}
After obtaining $F_{{pr}}$, we need to align the RGB modality feature $F_{{r}}$ with the prior multimodal feature $F_{{pr}}$ due to the large difference between them. For this purpose, we further design a homologous-modality feature alignment (HMFA) module, as shown in Figure~\ref{fig_HMFA}. 
We measure the similarity of these two features as an explicit alignment, and perform an implicit alignment via a Swin Transformer Alignment (STA) block which is inspired by \cite{liu2021swin}. 
The main purpose behind it is to use the self-attention mechanism to learn the alignment relationship between $F_{{r}}$ and $F_{{pr}}$.

First, a HMFA module is composed of several STA blocks, and each STA block aligns the RGB modality with the prior homologous multimodal feature based on a multi-headed attention mechanism.
\begin{equation}
F_{{an}} = \mathcal{B}_{STA}(F_{{pr}},F_{{r}}),
\label{equ_FAN}
\end{equation}
where $F_{{an}}$ denotes the aligned output feature by the STA module. $\mathcal{B}_{STA}(\cdot)$ denotes a STA block, which contains two feature encoders including a window multimodal encoder (wmE) and a shifted window multimodal encoder (swmE). The STA blocks are organized by Eq.~(\ref{equ_Recursive1}). 
\begin{equation}
\resizebox{.85\hsize}{!}{$
F_{{an}}^{i+1} = \begin{cases}
\mathcal{B}_{sw}^{i}(Q=F_{{m}}^{i},K,V=\mathcal{F}_{conv}^{i}(F_{{pr}})) &i=1\\
\mathcal{B}_{sw}^{i}(Q=F_{{m}}^{i},K,V=\mathcal{F}_{conv}^{2i-1}(F_{{pr}})) 
&n \geq i>1
\end{cases}$},
\label{equ_Recursive1}
\end{equation}
where $\mathcal{{B}}_{w}^{i}(Q,K,V)$ and $\mathcal{{B}}_{sw}^{i}(Q,K,V)$ denote the \textit{i}-th \texttt{{wmE}} and \texttt{{swmE}}, respectively. 
$(Q,K,V)$ is the query, key, and value in the self-attention mechanism.  $F_{M}^{i}$ denotes the output of the \textit{i}-th \texttt{{wmE}}, which is defined by 
\begin{equation}
\resizebox{.85\hsize}{!}{$
F_{{m}}^{i} = \begin{cases}
\mathcal{B}_{w}^{i}(Q=F_{{r}}, K,V = F_{{pr}})) &i=1\\
\mathcal{B}_{w}^{i}(Q = F_{{an}}^{i-1}, K,V = \mathcal{F}_{conv}^{2i-2}(F_{{pr}})))
&n \geq i>1
\end{cases}$},
\label{equ_Recursive2}
\end{equation}
where $\mathcal F_{conv}^{i}(\cdot)$ denotes the \textit{i}-th convolution block, and each convolution block consists of a two-layer convolution.

The difference between \texttt{{wmE}} and \texttt{{swmE}} is mainly in the shifted window mechanism for the interaction between different windows, which are used interchangeably in the STA module. 
The benefit of such a structure strengthens the memory of the deep network by repeatedly connecting the prior multimodal feature via the self-attention mechanism, gradually capturing the global information of multimodal features, learning the homologous multimodal interrelationships, and aligning the RGB and prior modalities.

\section{Experimental Validations}\label{sec:experiment}
\subsection{Experimental Protocols}
\subsubsection{\textbf{Dataset description}.} 
\label{subsec:experimental_dataset}
We trained hmJND-Net based on the latest benchmark dataset \cite{shen2020just}, which covers various image contents, including outdoor, indoor, landscape, nature, people, objects and buildings.
The dataset consists of 202 high-definition original images with the size of 1920$\times$1080 and 7878 redundancy-removed images by VVC.
Each original image has 39 encoded versions with different redundancy-removed levels, which is selected by the subjective experiments.
\begin{figure}[!t]
    \centering
    \begin{subfigure}[b]{0.09\textwidth}
        \includegraphics[width=\textwidth]{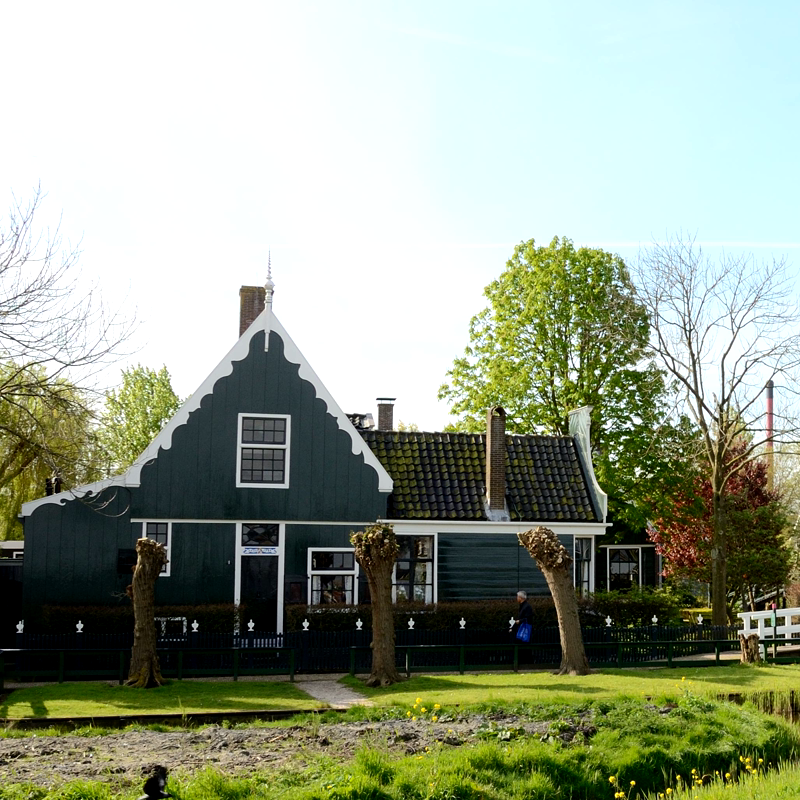}
        \captionsetup{font={scriptsize}}
        \caption{\textit{Original}}
        \label{fig_JNDmaps_a}
    \end{subfigure}
    \begin{subfigure}[b]{0.09\textwidth}
        \includegraphics[width=\textwidth]{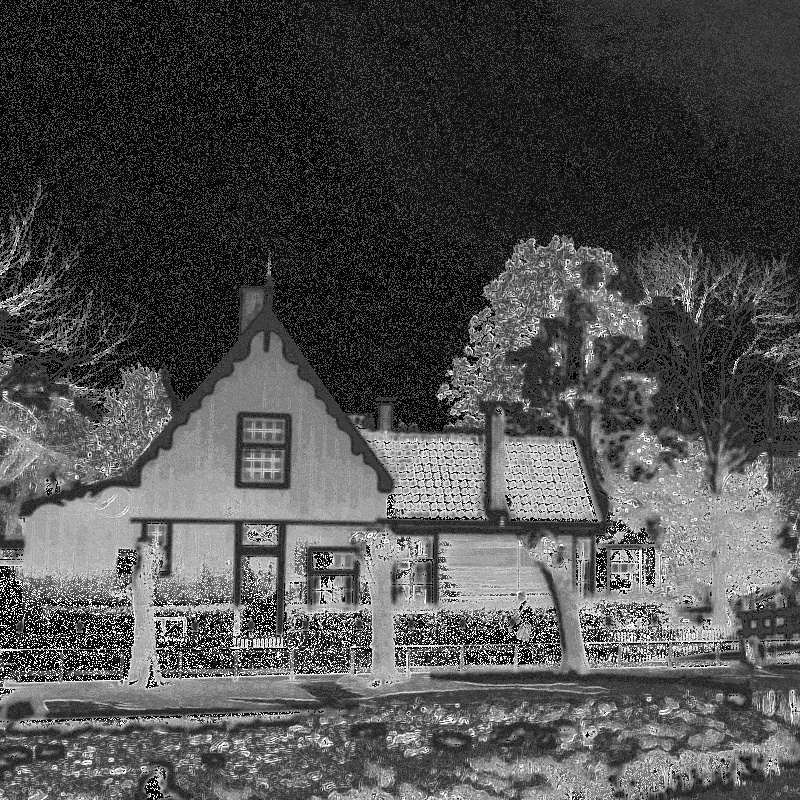}
        \captionsetup{font={scriptsize}}
        \caption{\textit{Yang2005}}
        \label{fig_JNDmaps_b}
    \end{subfigure}
    \begin{subfigure}[b]{0.09\textwidth}
        \includegraphics[width=\textwidth]{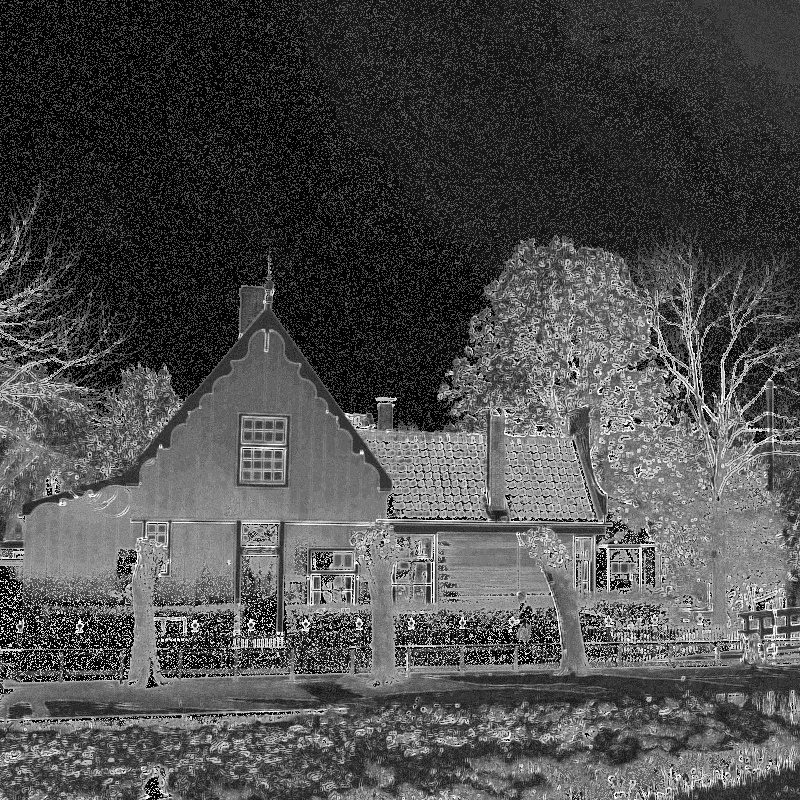}
        \captionsetup{font={scriptsize}}
        \caption{\textit{Liu2010}}
        \label{fig_JNDmaps_c}
    \end{subfigure}
    \begin{subfigure}[b]{0.09\textwidth}
        \includegraphics[width=\textwidth]{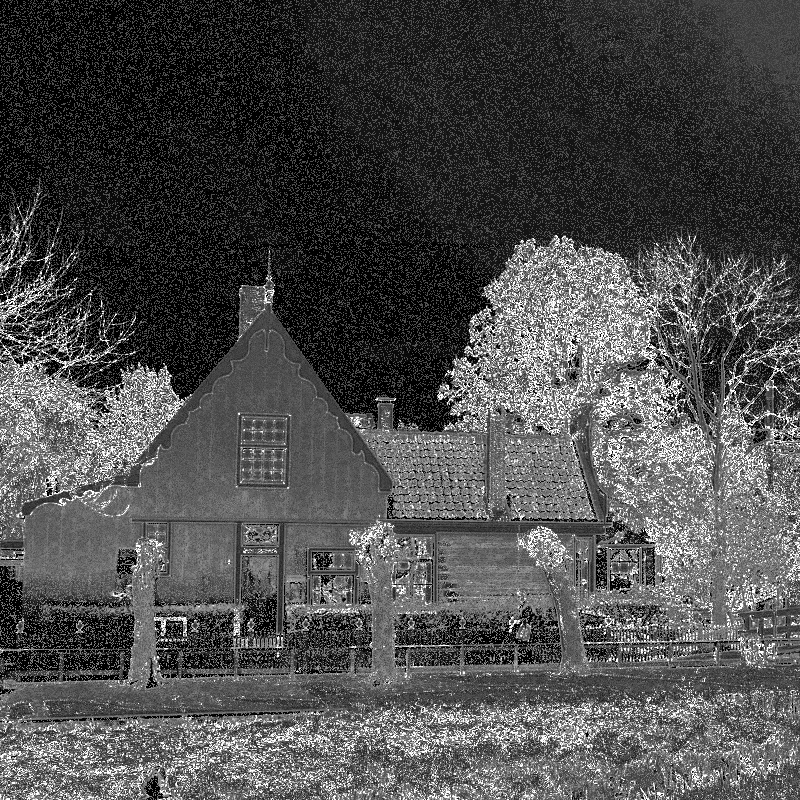}
        \captionsetup{font={scriptsize}}
        \caption{\textit{Wu2013}}
        \label{fig_JNDmaps_d}
    \end{subfigure}
    \begin{subfigure}[b]{0.09\textwidth}
        \includegraphics[width=\textwidth]{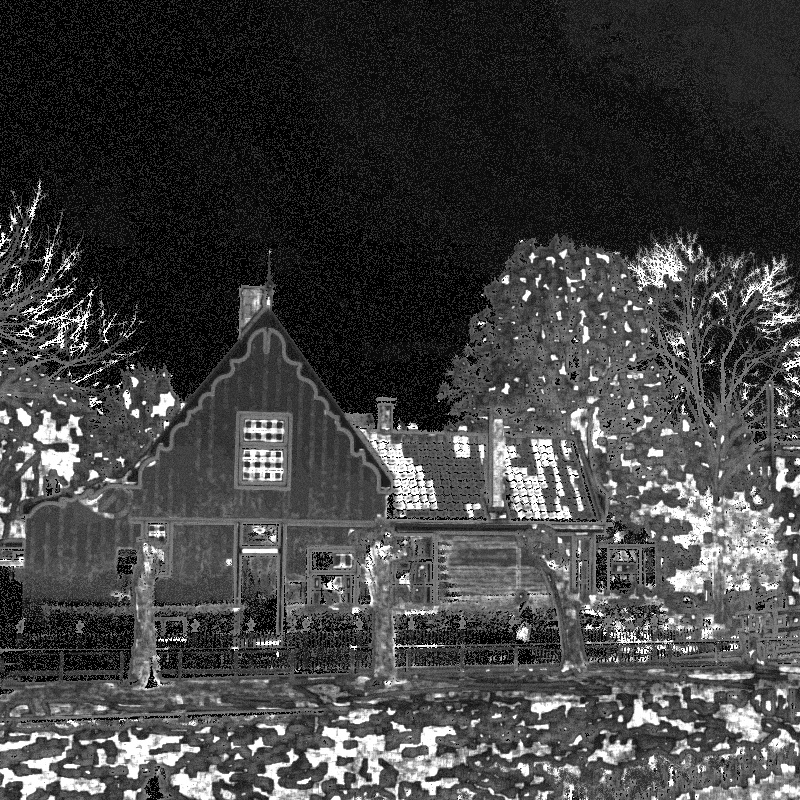}
        \captionsetup{font={scriptsize}}
        \caption{\textit{Wu2017}}
        \label{fig_JNDmaps_e}
    \end{subfigure}

    \begin{subfigure}[b]{0.09\textwidth}
        \includegraphics[width=\textwidth]{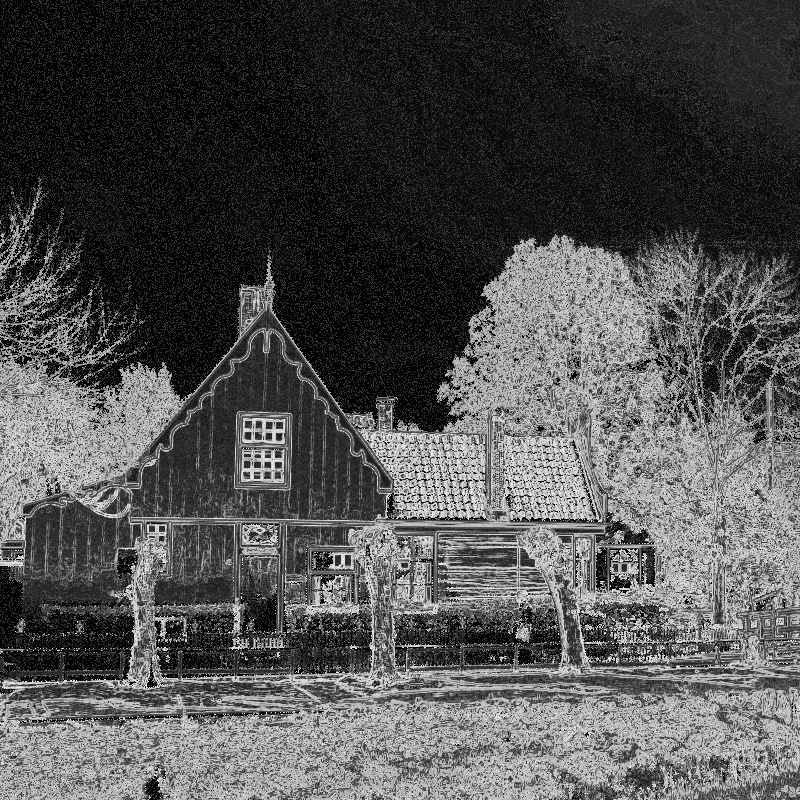}
        \captionsetup{font={scriptsize}}
        \caption{\textit{Chen2019}}
        \label{fig_JNDmaps_f}
    \end{subfigure}
    \begin{subfigure}[b]{0.09\textwidth}
        \includegraphics[width=\textwidth]{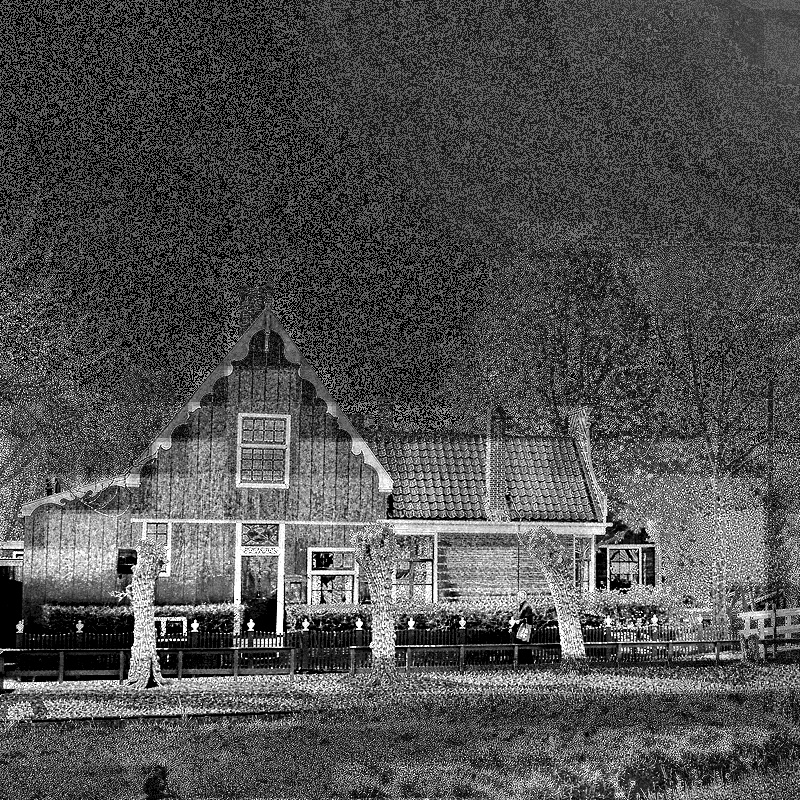}
        \captionsetup{font={scriptsize}}
        \caption{\textit{Shen2020}}
        \label{fig_JNDmaps_g}
    \end{subfigure}
    \begin{subfigure}[b]{0.09\textwidth}
        \includegraphics[width=\textwidth]{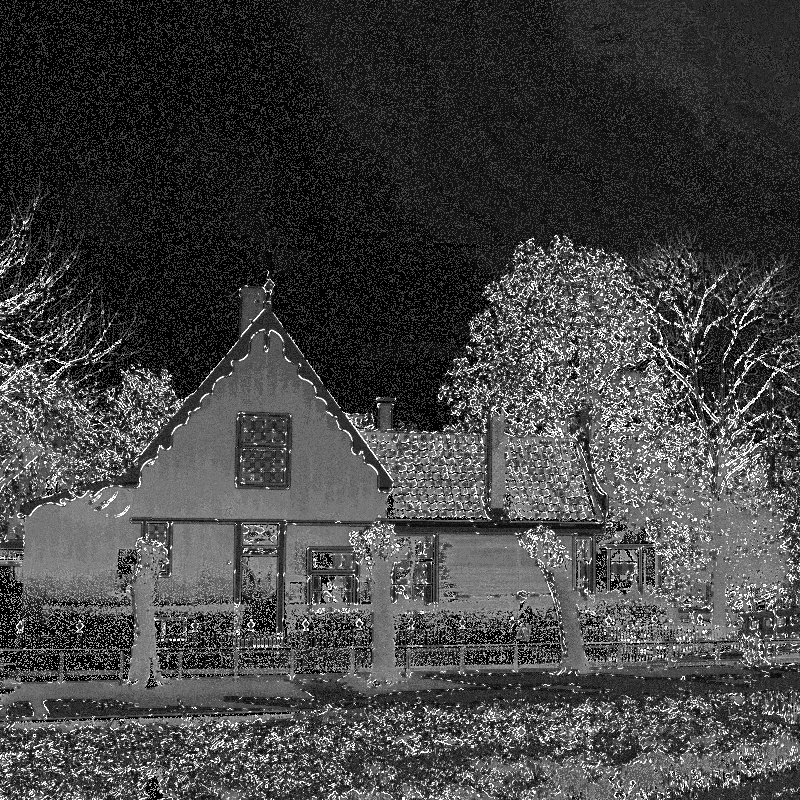}
        \captionsetup{font={scriptsize}}
        \caption{\textit{Wang2022}}
        \label{fig_JNDmaps_h}
    \end{subfigure}
    \begin{subfigure}[b]{0.09\textwidth}
        \includegraphics[width=\textwidth]{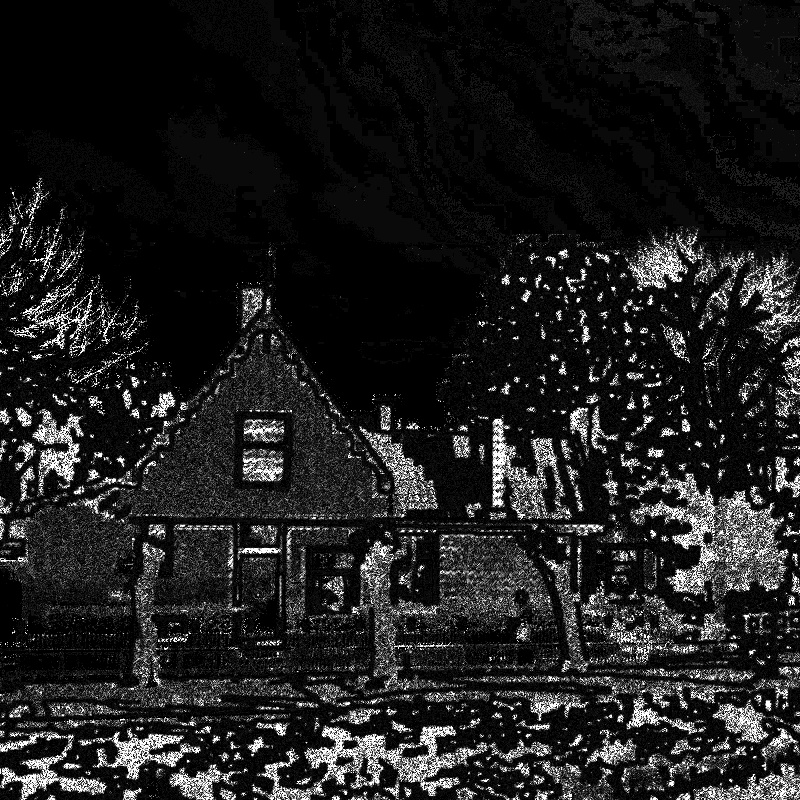}
        \captionsetup{font={scriptsize}}
        \caption{\textit{Jiang2022}}
        \label{fig_JNDmaps_i}
    \end{subfigure}
    \begin{subfigure}[b]{0.09\textwidth}
        \includegraphics[width=\textwidth]{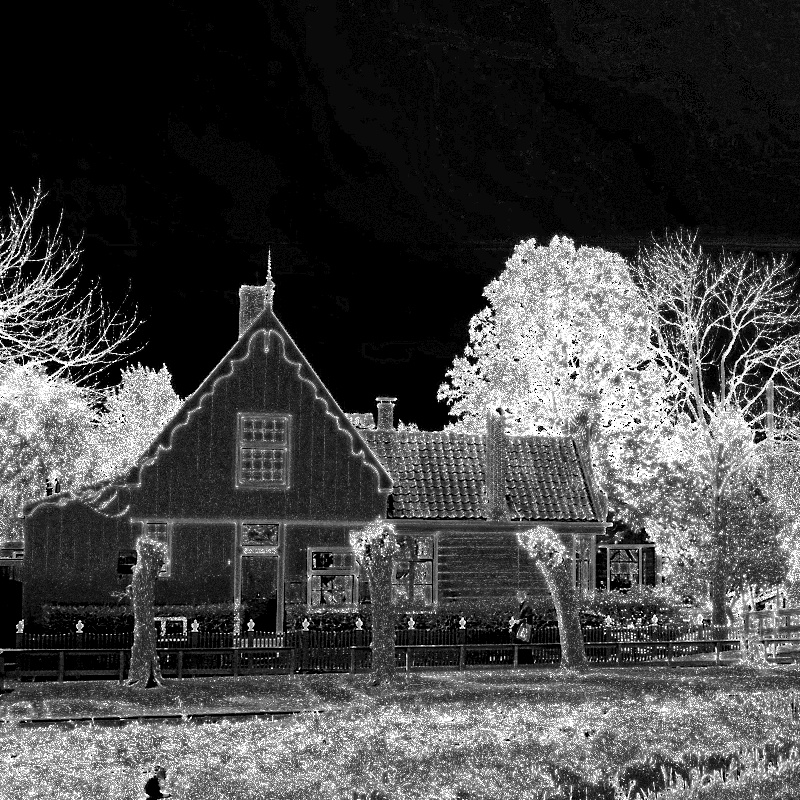}
        \captionsetup{font={scriptsize}}
        \caption{\textit{Ours}}
        \label{fig_JNDmaps_j}
    \end{subfigure}

    \caption{\textbf{Comparisons of the predicted visual redundancy for 9  representative approaches}. In {Figures (b)-(j)}, brighter pixels represent higher redundancies.}
\label{fig_JNDmaps}
\end{figure}

\begin{figure}[!t]
\centering
\includegraphics[width=\linewidth]{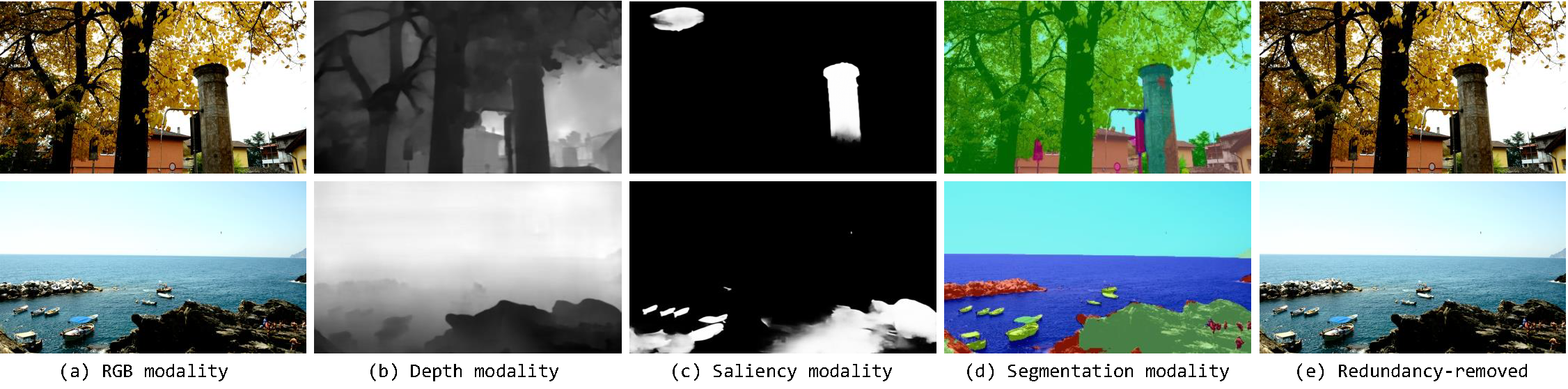}
\caption{\textbf{Examples of four modalities used in the experiments}. The right-most column is the ground-truth of $I_{rr}$.
}
\label{fig_datasets}
\end{figure}

The three modalities used in the experiments were obtained from the original images, with the depth modality generated by DPT-Hybrid \cite{ranftl2021vision}, the saliency modality generated by BASNet \cite{qin2019basnet}, and the segmentation modality generated by BEIT \cite{bao2021beit}. We provide some typical examples as shown in Figure~\ref{fig_datasets}.

We randomly select image pairs from the dataset as the training set, validating set, and testing set based on the ratio of \textbf{8:1:1}. 
To demonstrate the generalization capability, we also independently conduct the testing experiments of seven additional benchmark datasets, including 
\textit{CSIQ} \cite{larson2010most} (30 original natural-content images with the size of 512$\times$512), 
\textit{KADID-10K} \cite{lin2019kadid} (81 original natural-content images with the size of 512$\times$384), 
\textit{LIVE} \cite{sheikh2005live} (29 original natural-content images with various resolutions), 
\textit{TID2013} \cite{ponomarenko2015image} (25 original natural-content images with the size of 512$\times$384), 
\textit{SCID} \cite{yang2015perceptual} (40 original screen-content images with the size of 1280$\times$720), 
\textit{SIQAD} \cite{ni2017esim} (20 original screen-content images with various  resolutions), 
and \textit{MCL-JCI} \cite{jin2016statistical} (50 original natural-content images with the size of  1920$\times$1080).

\subsubsection{\textbf{Evaluation metrics}.} A better JND approach is able to hide more noise. To evaluate the noise tolerance performance, we randomly eject noise into each pixel guided by the predicted visibility threshold.
\begin{equation}
I_{con} =  I_{ori} + \alpha \times \gamma \times I_{vt},
\label{equ_injected}
\end{equation}
where $I_{con}$ denotes the JND-contaminated image and $I_{ori}$ denotes the original image. $\alpha$ denotes an adjustment factor used to adjust the noise level. $\gamma$ represents a random matrix with the same size as $I_{ori}$, and its value is 1 or -1. 
$I_{vt}$ denotes the predicted visibility threshold by the corresponding method. In the experiments. $I_{vt}$ represent the visibility threshold, which is the difference between $I_{ori}$ and $I_{rr}$.

In the experiments, we adjust $\alpha$ in Eq. (\ref{equ_injected}) to ensure all JND-contaminated images at the same noise level (\textit{e.g.}, mean square error (MSE)=100). To better demonstrate the comparison performance, we evaluate the prediction accuracy of visibility threshold at the image level.  Specifically, we use the Peak Signal-to-Noise Ratio (PSNR) and the Structural Similarity (SSIM) as the baseline indicators. 
We denote the PSNR result between the redundancy-removed image and the ground-truth image as $PSNR^{gt}_{rr}$, and the SSIM result as $SSIM^{gt}_{rr}$.

We employ the MS-SSIM and the mean opinion score (MOS) to measure the visual quality of contaminated images generated by different models \cite{zhai2020perceptual}. 
MS-SSIM is an objective full-reference image quality indicator with a score in the range of [0,1]. {$MS^{ori}_{con}$} denotes the MS-SSIM between the JND-contaminated image and the original one. Correspondingly, MOS evaluates the visual quality difference, denoted as $MOS^{ori}_{con}$, which is obtained by a subjective experiment.

In summary, $PSNR^{gt}_{rr}$ and $SSIM^{gt}_{rr}$ measure the forecasting accuracy of redundancy-removed  images. {$MS^{ori}_{con}$} and $MOS^{ori}_{con}$ measure the perceptual quality of JND-contaminated images. A higher value means a better result for all four metrics.

\subsubsection{\textbf{Subjective test}.}
The subjective experiment is conducted mainly based on ITU-R BT.500-13 standard. We invite 20 subjects to participate in the subjective test, 12 males and 8 females, aging from 18 to 40. 
Two images are juxtaposed on a 65-inch monitor, where one of them is the original image and the other one is the JND-contaminated image. The subjective progress is the same as in \cite{shen2020just}. After obtaining the subjective results, we employ a statistical-based outlier rejection method suggested in \cite{shen2020just} to examine the consistency of raw results in the experiments. After removing the outlier data, the average result is considered as the final MOS score in the following section.

\subsubsection{\textbf{Implementation details}.} hmJND-Net has been implemented on \textit{PyTorch} with all weights initialized by the \textit{truncated normal initializer}, and the \textit{Adam} optimizer with the default parameters (\textit{e.g.}, $\beta_{1}$ = 0.9 and $\beta_{2}$ = 0.999). For the HMFA module, we have executed 3 sets of STA blocks. 
We have trained the hmJND-Net with a mini-batch size of 16 for 200 epochs on an \textit{Nvidia Tesla A100} GPU, which takes about 10 hours. The initial learning rate is 1e-4, which will linearly decay to 0 from the 200-\textit{th} epoch. 
Due to the large image size (1920$\times$1080) in the dataset, an adaptive partition is performed on all input images, with the partition size of 224$\times$224. Apart from the above operations, we do not use any additional special data augmentation methods.

\subsection{Ablation Study}
To verify the necessity of the introduced prior modalities and the validity of the proposed multimodal fusion and alignment modules, we perform the ablation study on the benchmark dataset  \cite{shen2020just}. In Table \ref{tab:ablation_modality} and Table~\ref{tab:ablation_module}, the symbols ``\Checkmark''  and ``\XSolidBrush'' mean selected and unselected, respectively.
\begin{table}[!t]
    \centering   
    \renewcommand{\arraystretch}{0.99}  
    \caption{Ablation study of three additional modalities on the proposed hmJND-Net.}
    \label{tab:ablation_modality}
    \scalebox{0.85}{
    \begin{tabular}{l|c|c|cr}
    \hline
    \hline
        Saliency & Depth & Segmentation & $PSNR^{gt}_{rr}[+]$ & $SSIM^{gt}_{rr}[+]$\\
    \hline
    \hline
        \XSolidBrush & \XSolidBrush & \XSolidBrush & 33.8645 & 0.9571 \\
        \Checkmark & \XSolidBrush & \XSolidBrush & 34.3585 & 0.9634 \\
        \XSolidBrush & \Checkmark & \XSolidBrush & 34.3719 & 0.9637 \\
        \XSolidBrush & \XSolidBrush & \Checkmark & 34.1246 & 0.9631 \\
        \Checkmark & \Checkmark & \Checkmark & \textbf{34.5038} & \textbf{0.9652} \\
    \hline
    \hline
    \end{tabular}}
\end{table}

\subsubsection{\textbf{Effects of three modalities}.} To verify the necessity of the introduced three modalities, we have conducted five related  experiments as tabulated in Table \ref{tab:ablation_modality}. 
In the experiments, the substitution of a certain modality is done by repeating the enabled modality. The experimental results show that the best results are achieved when three modalities are used simultaneously, which verify the necessity of the introduced modalities.

\subsubsection{\textbf{Effects of HMPF and HMFA}.} To validate the effectiveness of our HMPF and HMFA modules, we have also conducted four additional experiments as provided in Table~\ref{tab:ablation_module}. 
In the experiments, the alternative to HMPF is to simply concatenate all prior modalities together, and then use a three-layer 3$\times$3 convolution and a one-layer 1$\times$1 convolution to reduce the number of features. 
\begin{table}[!t]
    \centering
    \caption{Ablation study of the proposed HMPF and HMFA modules.}
    \label{tab:ablation_module}
    \renewcommand{\arraystretch}{0.99} 
   \scalebox{0.85}{ 
   \setlength{\tabcolsep}{3.5mm}{ 
    \begin{tabular}{l|c|cr}
    \hline
    \hline
        {HMPF} & {HMFA} & $PSNR^{gt}_{rr}[+]$ & $SSIM^{gt}_{rr}[+]$\\
    \hline
    \hline
        \XSolidBrush & \XSolidBrush & 33.8348 & 0.9594 \\
        \XSolidBrush & \Checkmark & 34.2866 & 0.9633 \\
        \Checkmark & \XSolidBrush & 34.0914 & 0.9608 \\
        \Checkmark & \Checkmark & \textbf{34.5038} & \textbf{0.9652} \\
    \hline
    \hline
    \end{tabular}}}
\end{table}

The alternative to HMFA is to first concatenate the prior {multimodal} features and the RGB modality together, and then use the channel attention mechanism based on SENet and a one-layer 1$\times$1 convolution to align them. The experimental results show that the best performance can be achieved when both two modules are enabled simultaneously, which verify the effectiveness of the proposed HMPF and HMFA modules.
\begin{table*}[!t]
    \centering
    \caption{The average $MS^{ori}_{con}$ results between 9 representative  methods on 8 benchmark datasets. The best result in each row is highlighted in \textbf{bold}.}
    \label{tab:SSIM_results}
    \renewcommand{\arraystretch}{0.99} 
    \scalebox{0.90}{
    \begin{tabular}{r|c|c|c|c|c|c|c|c||r}
    \hline
    \hline
    \diagbox{Method}{Dataset} & \textit{CSIQ} & \textit{KADID-10K} & \textit{LIVE} & \textit{TID2013} & \textit{SCID} & \textit{SIQAD} & \textit{MCL-JCI} & \textit{SHEN2020} & \textbf{Average} \\
    \hline
    \hline
\textit{Yang2005} \cite{yang2005just}            & 0.9591 & 0.9513 & 0.9543 & 0.9519 & 0.9525 & 0.9686 & 0.9352 & 0.9510 & {0.9530} \\
\textit{Liu2010} \cite{liu2010just}             & 0.9629 & 0.9547 & 0.9576 & 0.9548 & 0.9558 & 0.9712 & 0.9370 & 0.9495 & {0.9554} \\
\textit{Wu2013} \cite{wu2013just}              & 0.9598 & 0.9541 & 0.9551 & 0.9520 & 0.9676 & 0.9831 & 0.9365 & 0.9504 & {0.9573} \\
\textit{Wu2017} \cite{wu2017enhanced}          & 0.9647 & 0.9617 & 0.9629 & 0.9599 & 0.9650 & 0.9761 & 0.9490 & 0.9593 & {0.9623} \\
\textit{Chen2019} \cite{chen2019asymmetric}      & 0.9690 & 0.9659 & 0.9658 & 0.9630 & 0.9691 & 0.9765 & 0.9567 & 0.9634 & {0.9662} \\
\textit{Shen2020} \cite{shen2020just}            & 0.9558 & 0.9454 & 0.9477 & 0.9452 & 0.9463 & 0.9609 & 0.9301 & 0.9418 & {0.9466} \\
\textit{Wang2022} \cite{wang2022perceptually}    & 0.9677 & 0.9613 & 0.9616 & 0.9586 & 0.9677 & 0.9820 & 0.9412 & 0.9581 & {0.9623} \\
\textit{Jiang2022} \cite{jiang2022towards}        & 0.9599 & 0.9689 & 0.9620 & 0.9591 & 0.9724 & 0.9772 & 0.9576 & 0.9627 & {0.9650} \\
\textit{Ours }  & \textbf{0.9709} & \textbf{0.9737} & \textbf{0.9695} & \textbf{0.9666} & \textbf{0.9795} & \textbf{0.9880} & \textbf{0.9584} & \textbf{0.9710} & \textbf{0.9722} \\
    \hline
    \hline
    \end{tabular}}
\end{table*}

\subsection{Overall Performance Comparisons}
\subsubsection{\textbf{Qualitative analysis}.} 
Figure~\ref{fig_JNDmaps} demonstrates a visual comparison of ``I05'' on SHEN2020 dataset, where Figure~\ref{fig_JNDmaps} (a) is the original image, and Figure~\ref{fig_JNDmaps} (b)-(j) provide the results of nine predicted visual redundancy, where brighter pixels mean higher redundancies. The detailed information for each method (\textit{e.g.}, \textit{Yang2005}) is provided in Table~\ref{tab:SSIM_results}.

As seen, existing representative methods may  predict high visibility thresholds on the salient objects or flat regions, which leads to worse visual quality.
The main reason is that smooth and flat regions are more sensitive, and a small change may attract the attention of HVS. 
Due to the joint guidance of the multimodal information, hmJND-Net tolerates more noise in texture regions (\textit{e.g.}, trees and grasses) which are not easily noticed by HVS, and tolerates less noise in smooth and flat sensitive regions. Thus, our hmJND-Net has a better redundancy prediction as shown in Figure~\ref{fig_JNDmaps} (j).

\subsubsection{\textbf{Quantitative analysis}.} 
The quantitative experiments are carried out on the eight widely-used image datasets as described in Sec. \ref{subsec:experimental_dataset}. In the experiments, the noise ejection level is MSE=100 for all testing images by Eq. \eqref{equ_injected}.

Table \ref{tab:SSIM_results} provides  the $MS^{ori}_{con}$ results of nine representative models on eight benchmark datasets. 
It can be seen that the proposed method achieves the highest average MS-SSIM results under the same noise level on all eight datasets.

In addition, we also conduct a subjective experiments under the same noise level (MSE=100) in terms of $MOS^{ori}_{con}$ on the SHEN2020 dataset. The average $MOS^{ori}_{con}$ results of each model (from top to bottom in Table \ref{tab:SSIM_results}) is -1.5071, -1.7214, -1.5929, -1.1429, -1.0572, -1.8964, -1.1842, -1.0833 and -0.5536, respectively. As seen, our hmJND-Net achieves the highest average MOS result.

\subsection{JND-guided Compression Performance} \label{sec:compression}
In this section, we have incorporated our hmJND-Net into the compression application, including the widely-used JPEG, and two state-of-the-art compression standards (\textit{e.g.}, HEVC and VVC).

\subsubsection{\textbf{JND-guided JPEG Compression}.}
In JND-guided JPEG compression, one widely-used manner is to remove visual redundancy of an input image before encoding, which can save bit-rate while maintaining the same visual quality. Specifically, each 8$\times$8 block is processed by the predicted visibility threshold as follows:
\begin{equation}
\resizebox{.85\hsize}{!}{$
\widehat{I}_{ori}(\mathbf{\mathbf{p}})=\left\{
            \begin{array}{ll}
            \overline{I}_{b}, & if\;|I_{ori}(\mathbf{p}) - \overline{I}_{b}| \le  I_{vt}(\mathbf{p}) ,\\
            I_{ori}(\mathbf{p})+I_{vt}(\mathbf{p}),  & if\;I_{ori}(\mathbf{p}) - \overline{I}_{b}<  -I_{vt}(\mathbf{p}),\\ 
            I_{ori}(\mathbf{\mathbf{p}})-I_{vt}(\mathbf{p}),  & otherwise.  
            \end{array}
            \right.$},
\label{equ_JPEG_Compression}
\end{equation}
where $I_{ori}(\mathbf{p})$ denotes a pixel magnitude at the position $\mathbf{p}$, $I_{vt}(\mathbf{p})$ denotes the related visual threshold value, $\overline{I}_{b}$ denotes the mean luminance value of the current 8$\times$8 block, and $\widehat{I}_{ori}(p)$ denotes the pre-processed output.

\subsubsection{\textbf{JND-Guided HEVC/VVC Compression}.}
In HEVC or VVC standards, we first obtain the JND-guided residual value, and the residual block is then encoded by the HEVC or VVC encoder. Specifically, we compute the JND-guided residual value as follows: 
\begin{equation}
\resizebox{.85\hsize}{!}{$
\widehat{R}(\mathbf{p}) =\left\{
            \begin{array}{ll}
            0, \qquad\qquad\qquad\quad ~~~~~~~~if\;|R(\mathbf{p})| \leq  I_{vt}(\mathbf{p})\;and\; \sigma^2(\mathbf{p}) \; > \sigma^{2},\\ 
            \mathbf{\min}(R(\mathbf{p}) \times \dfrac{|I_{vt}(\mathbf{p})|} {\sigma^2}, \; R(\mathbf{p}) + I_{vt}(\mathbf{p})),  \; else \; if \; R(\mathbf{p}) < 0,\\
            \mathbf{\max}(R(\mathbf{p}) \times \dfrac{|I_{vt}(\mathbf{p})|} {\sigma^2}, \; R(\mathbf{p}) - I_{vt}(\mathbf{p})),  \; otherwise.  
            \end{array}
            \right.$},
\label{equ_HEVC_Compression}
\end{equation}
where $\sigma^2(\mathbf{p})$ and $\sigma^2$ denote the related variances of a local range and the encoding block at the position $\mathbf{p}$, $R(\mathbf{p})$ denotes the residual value obtained by HEVC or VVC,  and $\widehat{R}(\mathbf{p})$ denotes the processed prediction residual value.
\begin{figure}[!t]
	\centering
	\setlength{\belowcaptionskip}{-0.1cm}
    \includegraphics[width=0.99\linewidth, height=0.65\linewidth]{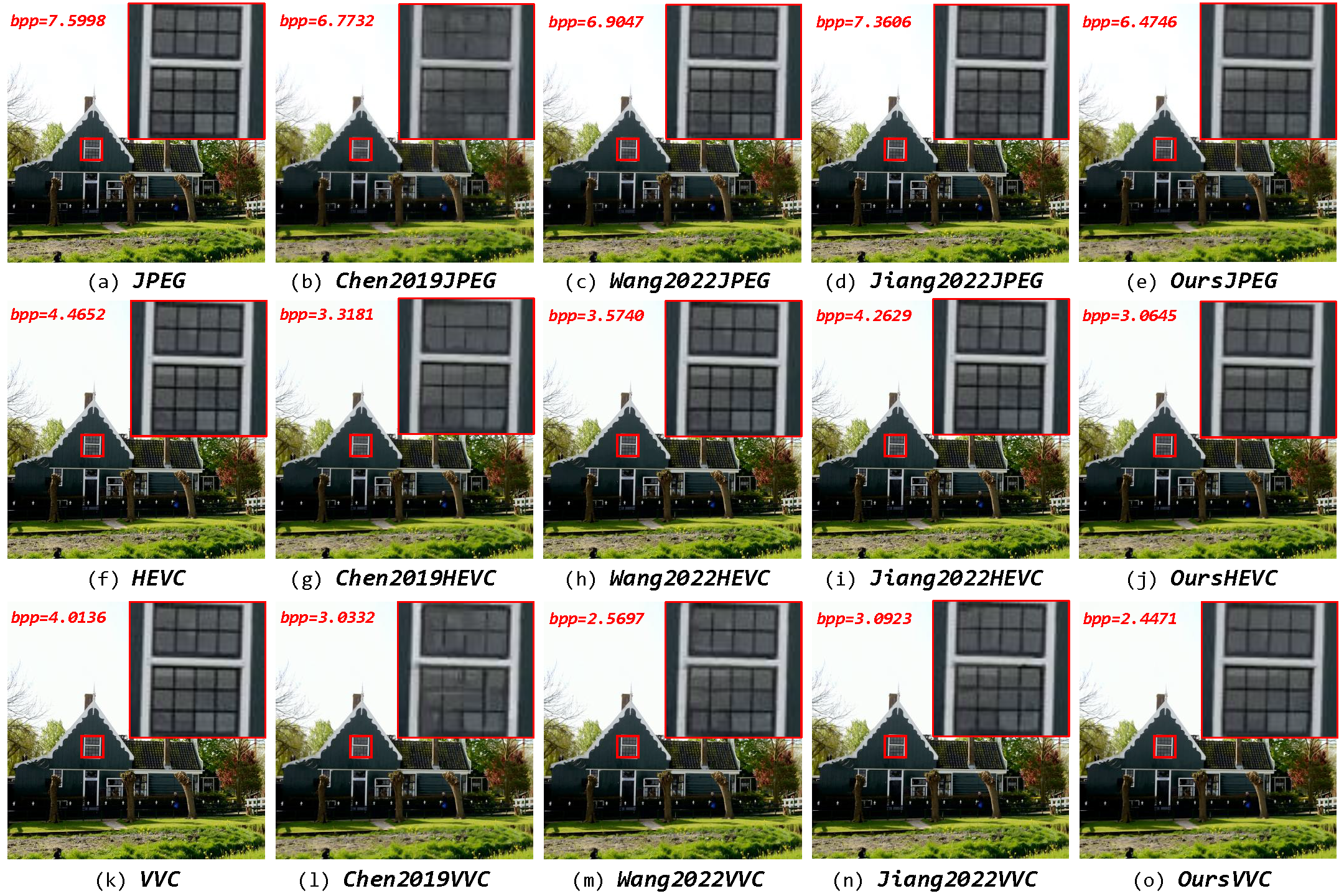}
	\caption{\textbf{Visual comparisons of four state-of-the-art JND-guided methods on three different encoding platforms.} }
\label{fig_Compression}
\end{figure}

\subsubsection{\textbf{Compression Performance Comparisons}.}
We have conducted the compression experiments on \textit{SHEN2020}, and four representative methods are compared in terms of bits-per-pixel (bpp). 
For example, the average bpp results of JPEG lossless, \cite{chen2019asymmetric} (\textit{Chen19JPEG}), \cite{wang2022perceptually} (\textit{Wang22JPEG}), \cite{jiang2022towards} (\textit{Jiang22JPEG}), and our hmJND-Net (\textit{OursJPEG}) are 4.3031, 3.4440, 3.5604, 4.0643, and 3.3771 on the JPEG encoder. 
The average bpp results of \textit{HEVC} lossless, \textit{Chen19HEVC}, \textit{Wang22HEVC}, \textit{Jiang22HEVC}, and \textit{OursHEVC} are 5.0411, 3.6557, 3.6366,  4.2253, and 3.4144 on the HEVC encoder. 
The average bpp results of \textit{VVC} lossless, \textit{Chen19VVC}, \textit{Wang22VVC}, \textit{Jiang22VVC}, and \textit{OursVVC} are 4.5304, 2.9958, 2.7764, 3.0934, and 2.6328 on the VVC encoder. 
With the similar visual quality, our hmJND-guided coding method saves the average bit-rate by $32.3809\%$.

In addition, Figure~\ref{fig_Compression} shows the visual compression performance comparisons of four methods on ``I05''.  As seen, our hmJND-Net-guided compression results provide nearly the same visual quality as  the lossless coding schemes, but our hmJND-guided approach saves the bit-rate by $28.4016\%$ on average.

\section{CONCLUSION}\label{sec:conclusion}
In this paper, we present an end-to-end visual redundancy forecasting network based on the homologous multimodal learning. Specifically, we consider three prior modalities to enrich the training data from different perspectives.
To effectively utilize the multimodal information, a homologous-modality prior fusion (HMPF) module is developed to fuse the prior modalities feature based on summation enhancement and subtractive offset.
In addition, we explore a homologous-modality feature alignment (HMFA)  module to align the intermediate features from different domains based on a self-attention driven encoder-decoder module. 
Experimental results on eight different benchmark datasets demonstrate the superiority of the method qualitatively and quantitatively. 
We believe that the proposed  multimodal-driven approach will help to improve the quality of service for the image compression and processing  systems.

\bibliography{dmmJND.bib}

\end{document}